\documentclass[letterpaper]{article} 
\usepackage{aaai2026}   
\usepackage{times}  
\usepackage{helvet}  
\usepackage{courier}  
\usepackage[hyphens]{url}  
\usepackage{graphicx} 
\usepackage{natbib}  
\usepackage{caption} 
\usepackage{algorithm}
\usepackage{algorithmic}
\usepackage{amsmath}
\usepackage{amssymb}
\usepackage{booktabs}
\usepackage{adjustbox}
\usepackage{cite}
\pdfoutput=1
\usepackage{newfloat}
\usepackage{listings}
\DeclareCaptionStyle{ruled}{labelfont=normalfont,labelsep=colon,strut=off} 
\floatstyle{ruled}
\newfloat{listing}{tb}{lst}{}
\floatname{listing}{Listing}
\title{GROVER: Graph-guided Representation of Omics and Vision with Expert Regulation for Adaptive Spatial Multi-omics Fusion}

\author{
Yongjun Xiao\textsuperscript{1,2}\equalcontrib, 
Dian Meng\textsuperscript{3}\equalcontrib,\\ 
Xinlei Huang\textsuperscript{1}, 
Yanran Liu\textsuperscript{1}, 
Shiwei Ruan\textsuperscript{1}, 
Ziyue Qiao\textsuperscript{1}, 
Xubin Zheng\textsuperscript{1}\thanks{Corresponding author.}
}

\affiliations{
    \textsuperscript{\rm 1}Dongguan Key Laboratory for Intelligence and Information Technology, School of Computing and Information Technology, Great Bay University, Guangdong, China\\
    \textsuperscript{\rm 2}College of Information Engineering, Sichuan Agricultural University, Ya’an, China\\
    \textsuperscript{\rm 3}Department of Biomedical Informatics, Yong Loo Lin School of Medicine, National University of Singapore, Singapore\\
    xbzheng@gbu.edu.cn

}

\usepackage{bibentry}

\begin{document}

\maketitle

\begin{abstract}
Effectively modeling multimodal spatial omics data is critical for understanding tissue complexity and underlying biological mechanisms. While spatial transcriptomics, proteomics, and epigenomics capture molecular features, they lack pathological morphological context. Integrating these omics with histopathological images is thus critical for comprehensive disease tissue analysis. However, substantial heterogeneity across omics, imaging, and spatial modalities poses significant challenges. Naive fusion of semantically distinct sources often leads to ambiguous representations. Additionally, the resolution mismatch between high-resolution histology images and lower-resolution sequencing spots complicates spatial alignment. Biological perturbations during sample preparation further distort modality-specific signals, hindering accurate integration. To address these challenges, we propose \textbf{G}raph-guided \textbf{R}epresentation of \textbf{O}mics and \textbf{V}ision with \textbf{E}xpert \textbf{R}egulation for Adaptive Spatial Multi-omics Fusion (\textbf{GROVER}), a novel framework for adaptive integration of spatial multi-omics data. GROVER leverages a Graph Convolutional Network encoder based on Kolmogorov–Arnold Networks to capture the nonlinear dependencies between each modality and its associated spatial structure, thereby producing expressive, modality-specific embeddings. To align these representations, we introduce a spot-feature-pair contrastive learning strategy that explicitly optimizes the correspondence across modalities at each spot. Furthermore, we design a dynamic expert routing mechanism that adaptively selects informative modalities for each spot while suppressing noisy or low-quality inputs. Experiments on real-world spatial omics datasets demonstrate that GROVER outperforms state-of-the-art baselines, providing a robust and reliable solution for multimodal integration.

\end{abstract}

\begin{links}
    \link{Code}{https://github.com/Xubin-s-Lab/GROVER}
\end{links}

\section{Introduction}
Spatially resolved transcriptomics and spatial proteomics were recognized as the Methods of the Year by Nature in 2021 \cite{marx2021method} and 2024 \cite{karimi2024method}, respectively. These advancements extend single-cell analyses of gene expression and surface protein abundance into the spatial domain, offering unprecedented insight into tissue organization. More recently, spatially resolved multimodal omics—encompassing transcriptomics, proteomics, and high-resolution histological imaging—have emerged as a powerful paradigm for integrative analysis, enabling a comprehensive understanding of gene regulation and the tissue microenvironment within their native spatial context \cite{coleman2025resolving}. The main challenge in multimodal spatial omics analysis is the effective integration of features from diverse modalities to generate coherent low-dimensional representations that facilitate downstream tasks, such as spatial domain identification via clustering.

\vspace{-0.3em}

Most recent approaches primarily focus on integrating transcriptomic and proteomic data \cite{meng2024scmmae}, while overlooking the valuable structural context provided by histological images. For instance, SpatialGlue \cite{long2024deciphering} employs cross-modal attention to fuse transcriptomic and proteomic features, and PRAGA \cite{huang2025praga} integrates spatial multi-omics data with adaptive graph structures and dynamic prototype contrastive learning. MISO \cite{coleman2025resolving} represents a recent advancement by incorporating histology images into a multimodal integration pipeline through outer-product interactions. Moreover, these methods tend to treat all modalities equally across spatial locations, ignoring substantial variations in data quality. In practice, spatial omics data are often compromised by noise from both technical limitations (e.g., dropout events in single-cell sequencing) \cite{ge2025deep} and biological or experimental artifacts (e.g., tissue sectioning errors in histology) \cite{totty2025spotsweeper}. Such heterogeneity makes it critical to assess the reliability of each modality per spot. Yet, current integration frameworks are unable to adaptively weigh or filter unreliable features based on local signal quality, which may limit the accuracy and robustness of spatial analysis. To address this, we propose a multi-expert learning framework that performs modality-aware integration at the single-spot level, selectively emphasizing high-confidence signals and mitigating the impact of noise. Furthermore, due to the substantial semantic gap between omics data and histological images, as well as the many-to-many mapping between image patches and spatial spots, achieving accurate cross-modal alignment remains highly challenging. To overcome this, we introduce a contrastive learning strategy that encourages consistency between spatial and morphological representations in a weakly paired setting.

In this paper, we propose a novel spatially resolved multi-omics framework, \textbf{G}raph-guided \textbf{R}epresentation of \textbf{O}mics and \textbf{V}ision with \textbf{E}xpert \textbf{R}egulation for Adaptive Spatial Multi-omics Fusion (GROVER). GROVER employs a spot-feature-pair based contrastive learning approach to integrate features from three modalities, guided by graph-based spatial relationships and structural information derived from histological images. Inspired by the Mixture-of-Experts (MoE) paradigm, we design a multi-expert model to adaptively filter and fuse modality-specific features at the spot level. Extensive qualitative and quantitative results demonstrate that GROVER excels at aggregating spatial multi-omics information into spot-type-resolvable representations by effectively integrating histological image features and performing adaptive feature fusion through a multi-expert framework. Our contributions are summarized as follows:

\begin{itemize}
    \item We propose a novel spatially resolved multi-omics framework, \textbf{GROVER}, which adaptively integrates transcriptomic, proteomic, and histological modalities at the single-spot level. We introduce a spot-feature-pair based contrastive learning strategy to bridge the semantic gap between omics data and histological images, enabling accurate cross-modal alignment.
    \item We design a modality-specific multi-expert architecture with gated routing to adaptively weigh heterogeneous signals, enhancing both robustness and interpretability.
    \item We develop a hybrid graph encoder, the Graph Convolutional Network based on Kolmogorov–Arnold Networks (KAN-GCN), which enhances message passing with kernel-based nonlinear transformations for expressive, structure-aware representation learning.
\end{itemize}

\section{Related Works}
\subsection{Spatial Multi-omics Integration}
Recently, spatial multi-omics technologies have emerged as powerful tools for linking spatial context with molecular profiling, offering new opportunities to dissect cellular heterogeneity within the tissue microenvironment. SpatialGlue \cite{long2024deciphering} integrates spatial and omics information using a dual-attention graph neural network that captures cross-modal correspondence and spatial structure. COSMOS \cite{zhou2025cooperative} combines graph convolutional networks (GCNs) with a weighted nearest neighbor (WNN) framework, and further employs Deep Graph Infomax (DGI) and spatial regularization to fuse complementary features for downstream analysis. However, these approaches primarily focus on integrating molecular modalities and often neglect histological context. Due to the semantic disparity and alignment difficulty between tissue images and omics measurements, image information is rarely incorporated into the integration process.

\subsection{Integration of Image and Omics Data}
Histological images offer crucial morphological context that complements molecular data, making them a valuable addition in spatial omics \cite{chelebian2025combining}. Their integration enhances both biological insight and analytical depth. MISO encodes each modality independently using Multilayer Perceptron (MLP), models interactions via outer products, and concatenates results into a unified embedding. However, it treats all modalities equally across spatial spots, ignoring differences in quality, noise, or relevance.To address this, we propose a multi-expert framework inspired by mixture-of-experts \cite{jacobs1991adaptive}. It assigns adaptive weights to modality-specific experts at each spot, allowing the model to prioritize informative signals while suppressing noisy or irrelevant inputs, leading to more robust and precise integration.

\section{Method}
\subsection{Preliminaries}


Spatial multi-modal omics integration aims to jointly model molecular measurements along with their spatial context, thereby obtaining unified spot-level representations suitable for downstream analyses. Formally, given a spatial dataset comprising $N$ sequencing spots, we denote their spatial coordinates as:
\begin{equation}
S = \{(x_i, y_i)\}_{i=1}^{N},\quad (x_i, y_i)\in\mathbb{R}^{2}.
\end{equation}

Each spot is characterized by measurements from $M$ different modalities, typically including transcriptomic, proteomic, and histological imaging modalities. For each modality $m \in \{1,\dots,M\}$, the features associated with the spots are defined as:
\begin{equation}
F^{(m)} = \{f_i^{(m)}\in \mathbb{R}^{D_m}\}_{i=1}^{N},
\end{equation}
where $f_i^{(m)}$ is the preprocessed $D_m$-dimensional feature vector corresponding to spot $i$.

For example, $F^{(m)}$ could represent gene expression features reduced by principal component analysis (PCA), antibody-derived tag (ADT) protein abundances, or visual features extracted from histology images using a pre-trained pathology foundation model such as OmiCLIP \cite{chen2025visual}. Notably, our framework is highly modular and flexible, enabling the seamless incorporation of any state-of-the-art (SOTA) pathology foundation model, thus ensuring adaptability to future technological advances.

The goal of spatial multi-modal integration is to ues a function $\Phi$ that adaptively fuses all modality features, informed by spatial topology, into a unified low-dimensional representation $Z$:
\begin{equation}
Z = \Phi(F^{(1)}, \ldots, F^{(M)}, S).
\label{eq:Z}
\end{equation}

The integrated representation $Z$ is expected to preserve complementary biological signals across modalities while maintaining spatial coherence. This embedding can be used for various downstream tasks, such as spatial domain identification, cell type annotation, and tumor microenvironment characterization.

\begin{figure*}[t]
  \centering
  \includegraphics[width=\textwidth]{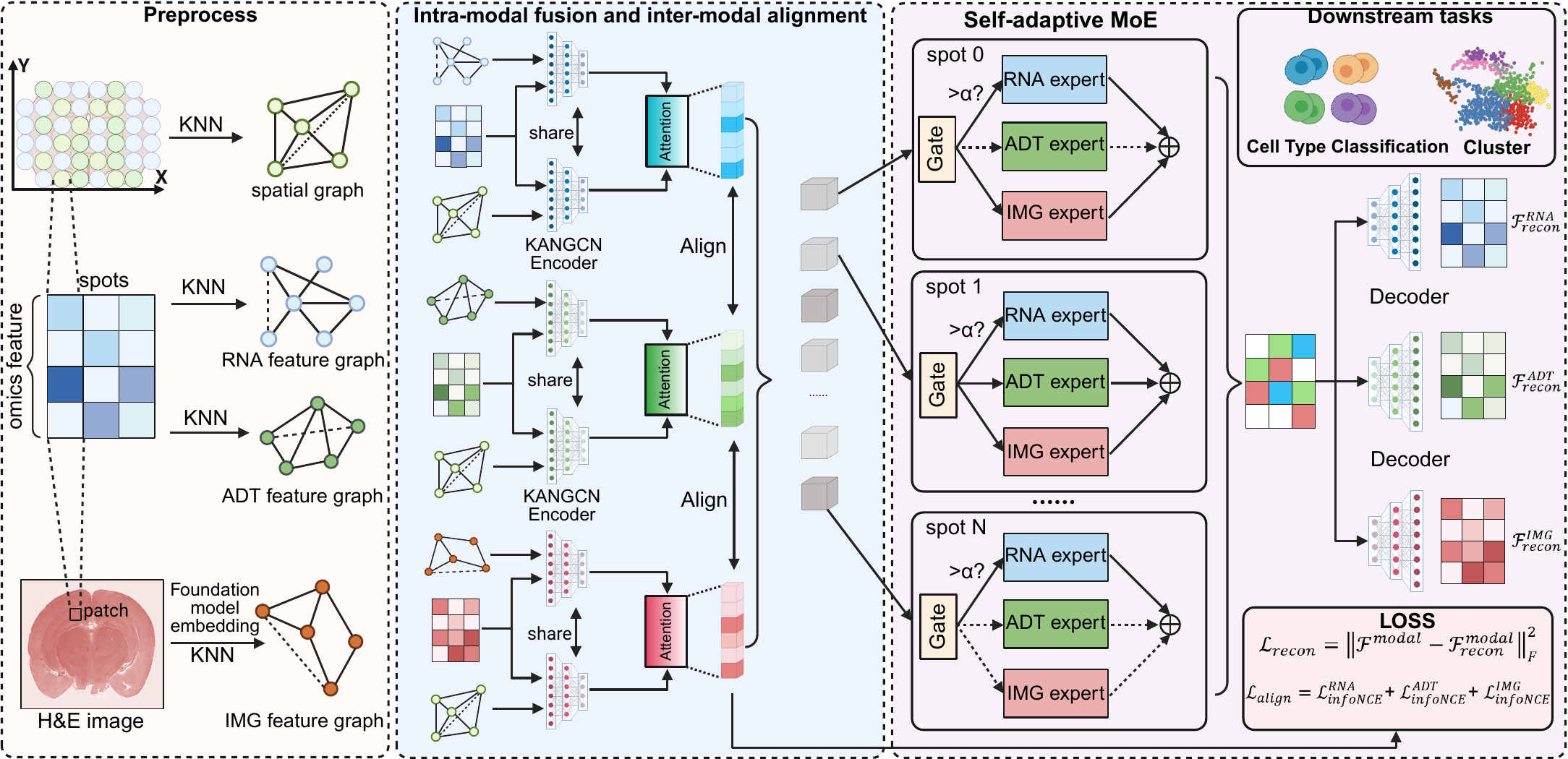}
  \caption{The framework of the proposed GROVER.GROVER encodes modality-specific feature graphs and spatial adjacency graphs using KAN-GCN, then applies attention-based weighted fusion to obtain integrated multimodal representations (RNA, protein, and image). A spot-feature-pair based contrastive learning aligns semantic information across modalities before feeding the embeddings into a self-adaptive Mixture-of-Experts model for fusion. The entire model is trained with modality-specific reconstruction losses and the spot-feature-pair contrastive loss.}
  \label{fig:grover}
\vspace{-1em}
\end{figure*}

\subsection{GROVER}

In this paper, we propose GROVER, a graph-guided architecture that models the function $\Phi$ in Eq.~(1), adaptively integrating transcriptomic, proteomic, and histological modalities at single-spot resolution into a unified representation.

Given $M$ modalities $\{F^{(m)}\}_{m=1}^M$, including omics features (e.g., RNA and protein) and histological image features, GROVER constructs two graphs per modality: a spatial graph $\mathcal{G}_S = (S, A_S)$ based on spot coordinates $S = \{(x_i, y_i)\}_{i=1}^N$, and a modality-specific feature graph $\mathcal{G}_F^{(m)} = (F^{(m)}, A_F^{(m)})$. Both $A_S$ and $A_F^{(m)}$ are built via K-Nearest Neighbor (KNN), where $A_S$ reflects physical proximity and $A_F^{(m)}$ captures feature similarity. These dual graphs jointly encode spatial and semantic structure.

For each modality, we fuse embeddings from the two graphs using a spatial-feature attention module:
\begin{equation}
\tilde{e}_i^{(m)} = \text{Attention}(e_i^{S}, e_i^{F,(m)}),
\label{eq:attention_fusion}
\end{equation}
where \(e_i^{S}\) and \(e_i^{F,(m)}\) are obtained via KAN-GCN on $\mathcal{G}_S$ and $\mathcal{G}_F^{(m)}$, respectively.

To reduce modality discrepancy, we apply spot-level contrastive learning to align the fused embeddings $\tilde{e}_i^{(m)}$. To address varying modality quality, an adaptive multi-expert model with dynamic gating is introduced. The final fused representation $Z$ is obtained via this gating mechanism. An overview of GROVER is shown in Figure 1, with the full pipeline summarized in Algorithm~1.

\subsection{Encoder based on the KAN-GCN}

To extract meaningful node representations from both spatial and modality-specific feature graphs, we adopt a graph convolutional encoder framework. Traditional Graph Convolutional Networks (GCNs) are widely used for learning node embeddings via message passing over graph structures \cite{kipf2016semi}.

Formally, given a graph $\mathcal{G} = (V, A)$ with $N$ nodes, where $V$ denotes the set of spatial spots and $A \in {0,1}^{N \times N}$ is the adjacency matrix representing the spatial or functional connectivity between spots, the graph is characterized by its node feature matrix $X \in \mathbb{R}^{N \times d_0}$, where $d_0$ is the input feature dimension. In our context, each node corresponds to a spatial spot, and its associated feature vector may include modality-specific measurements such as gene expression, protein abundance, or chromatin accessibility.

The layer-wise propagation rule of a standard GCN is defined as:
\begin{equation}
H^{(l+1)} = \sigma\left( \tilde{D}^{-1/2} \tilde{A} \tilde{D}^{-1/2} H^{(l)} W^{(l)} \right),
\end{equation}
where $H^{(0)} = X$, $W^{(l)} \in \mathbb{R}^{d_l \times d_{l+1}}$ is the learnable weight matrix at layer $l$, and $\tilde{A} = A + I$ is the adjacency matrix with added self-loops, with $\tilde{D}$ being the corresponding degree matrix. 
Here, $l = 0, 1, \dots, L-1$ denotes the layer index, and $L$ is the total number of encoder layers.

To enhance the expressive power of GCNs, we replace the fixed linear transformation with a nonlinear function modeled by a Kolmogorov--Arnold Network (KAN) \cite{liu2024kan}. Each KAN layer consists of a matrix of trainable univariate functions $\varphi_{q,p}^{(l)} : [0, 1] \rightarrow \mathbb{R}$, where $p = 1, \dots, d_l$ and $q = 1, \dots, d_{l+1}$. The transformed features are computed as:
\begin{equation}
\mathcal{F}^{(l)}(H^{(l)})_{i,q} = \sum_{p=1}^{d_l} \varphi_{q,p}^{(l)}(H^{(l)}_{i,p}),
\end{equation}
and the node update becomes:
\begin{equation}
H^{(l+1)} = \sigma\left( \hat{A} \cdot \mathcal{F}^{(l)}(H^{(l)}) \right).
\end{equation}

In GROVER, we apply a multi-layer KAN-GCN encoder to the spatial graph $\mathcal{G}_S$ and each modality-specific graph $\mathcal{G}_F^{(m)}$, and denote the final-layer outputs as $H_S^{(L)} \in \mathbb{R}^{N \times d_L}$ and $H_{F^{(m)}}^{(L)} \in \mathbb{R}^{N \times d_L}$, respectively. The node embeddings for spot $i$ are:
\begin{equation}
e_i^S = H_{S,i}^{(L)}, \quad e_i^{F^{(m)}} = H_{F^{(m)},i}^{(L)}.
\end{equation}

To adaptively fuse these two types of embeddings, we use a within-modality attention aggregation layer. Let $\mathbf{W} \in \mathbb{R}^{d_L \times d_{\text{att}}}$ and $\mathbf{b} \in \mathbb{R}^{d_{\text{att}}}$ be the shared linear transformation parameters, and let $\mathbf{q} \in \mathbb{R}^{d_{\text{att}}}$ be a learnable attention vector. Here, $d_L$ is the dimensionality of the encoder output, and $d_{\text{att}}$ is a hidden projection dimension used specifically for computing attention scores.

The scalar compatibility score for each embedding source $t \in \{S, F\}$ is computed as:
\begin{equation}
e_i^{(t)} = \mathbf{q}^\top \tanh\left( \mathbf{W} e_i^{(t)} + \mathbf{b} \right).
\end{equation}

We then apply a softmax to normalize the scores into attention weights:
\begin{equation}
\alpha_i^{(t)} = \frac{\exp(e_i^{(t)})}{\exp(e_i^{(S)}) + \exp(e_i^{(F)})}.
\end{equation}

Finally, the fused modality-aware embedding is obtained by weighted summation:
\begin{equation}
\tilde{e}_i^{(m)} = \alpha_i^{(S)} e_i^S + \alpha_i^{(F)} e_i^{F^{(m)}},
\end{equation}
where $\tilde{e}_i^{(m)} \in \mathbb{R}^{d_L}$ serves as the unified representation for downstream tasks.

\subsection{Spot-Feature-Pair Based Contrastive Learning}

The transcriptomic (RNA), proteomic (ADT), and histological (image) modalities differ significantly in terms of data distributions and biological semantic characteristics.
To address this challenge, we introduce a dual-alignment strategy that aligns both spatial and semantic information prior to modality fusion. For spatial alignment, we extract image patches centered at each spot's absolute coordinates on the tissue section and encode them using a pretrained vision foundation model to obtain location-aware image embeddings.
For semantic alignment, we implement a bi-directional masked contrastive learning framework that enforces semantic consistency across pairs of modalities while mitigating the impact of spurious negatives caused by biologically similar spots.

Specifically, for each modality \( m \in \{\text{RNA}, \text{ADT}, \text{Image}\} \), we obtain fused embeddings \(\tilde{E}^{(m)} = \{\tilde{e}_i^{(m)}\}_{i=1}^N\).
We first compute a cosine similarity matrix:
\begin{equation}
S^{(m)}_{i,j} = \mathrm{sim}(\tilde{e}_i^{(m)}, \tilde{e}_j^{(m)}),
\end{equation}
and define a binary similarity mask \(M^{(m)} \in \{0,1\}^{N \times N}\) as:
\begin{equation}
M^{(m)}_{i,j} =
\begin{cases}
0, & \text{if } S^{(m)}_{i,j} > \delta \text{ and } i \ne j, \\
1, & \text{otherwise},
\end{cases}
\end{equation}
where \(\delta\) is a threshold controlling the exclusion of highly similar negatives.
This mask removes semantically similar spots from the contrastive denominator to mitigate false negatives during training.

Using the mask, we define the masked InfoNCE loss for a modality pair \((m_1, m_2)\) as:
\begin{equation}
\ell_{\mathrm{masked}}\big( \tilde{E}^{(m_1)}, \tilde{E}^{(m_2)}, M^{(m_1)} \big) 
= -\frac{1}{N} \sum_{i=1}^N \log \, s_i,
\end{equation}

\begin{equation}
s_i = \frac{
\exp\big( \mathrm{sim}(\tilde{e}_i^{(m_1)}, \tilde{e}_i^{(m_2)}) / \tau \big)
}{
\sum_{j=1}^N M^{(m_1)}_{i,j} \exp\big( \mathrm{sim}(\tilde{e}_i^{(m_1)}, \tilde{e}_j^{(m_2)}) / \tau \big)
},
\end{equation}
where \(\tau\) is a temperature parameter and \(\mathrm{sim}(\cdot, \cdot)\) denotes the cosine similarity between normalized vectors.

To ensure symmetric alignment between modalities, we compute the masked InfoNCE loss \cite{oord2018representation} in both directions and average them:
\begin{equation}
\begin{aligned}
\mathcal{L}_{\mathrm{contrast}}^{m_1,m_2} = \frac{1}{2} \Big(
& \ell_{\mathrm{masked}}(\tilde{E}^{(m_1)}, \tilde{E}^{(m_2)}, M^{(m_1)}) \\
+ & \ell_{\mathrm{masked}}(\tilde{E}^{(m_2)}, \tilde{E}^{(m_1)}, M^{(m_2)})
\Big).
\end{aligned}
\label{eq:bidirectional_loss}
\end{equation}

We apply this contrastive loss to three modality pairs: RNA--ADT, RNA--Image, and ADT--Image. This bi-directional masked contrastive strategy enhances the alignment of shared biological semantics across modalities while avoiding over-penalization of semantically similar yet non-identical spots. We denote the aligned embeddings as \(\hat{e}_i^{(m)}\).

\subsection{Self-adaptive Mixture of Experts}

In practice, due to quality differences among modalities, naively integrating all modalities at the single-spot resolution can be suboptimal. To address this, we propose a self-adaptive Mixture of Experts (MoE) framework that dynamically adjusts each modality's contribution per spot.

Given the aligned embeddings \(\hat{e}_i^{(R)}\), \(\hat{e}_i^{(A)}\), and \(\hat{e}_i^{(I)}\) from RNA, protein, and image modalities respectively, we first compute an aggregated gating input by averaging:
\begin{equation}
x_i = \frac{1}{3} \left( \hat{e}_i^{(R)} + \hat{e}_i^{(A)} + \hat{e}_i^{(I)} \right).
\end{equation}

A gating network parameterized by \(W_{\mathrm{gate}} \in \mathbb{R}^{D \times 3}\) produces raw confidence scores \(\mathbf{g}_i = W_{\mathrm{gate}} x_i \in \mathbb{R}^3\), which are converted via softmax to normalized weights:
\begin{equation}
\beta_i^{(m)} = \frac{\exp(g_i^{(m)})}{\sum_{m'} \exp(g_i^{(m')})}, \quad m \in \{R, A, I\}.
\end{equation}

To filter out unreliable modalities, we apply a threshold \(\gamma\) to the gating scores. In our experiments, we set \(\gamma = 0.3\).
\begin{equation}
\tilde{\beta}_i^{(m)} = 
\begin{cases}
\beta_i^{(m)}, & \text{if } \beta_i^{(m)} \geq \gamma \\
0, & \text{otherwise}
\end{cases}.
\end{equation}

These filtered weights are renormalized to sum to one:
\begin{equation}
s_i^{(m)} = \frac{\tilde{\beta}_i^{(m)}}{\sum_{m'} \tilde{\beta}_i^{(m')} + \epsilon},
\end{equation}
where \(\epsilon = 10^{-6}\) is a small constant for numerical stability.

Each modality has a dedicated expert implemented as a feed-forward network:
\begin{equation}
h_i^{(m)} = \mathrm{FFN}^{(m)}(\hat{e}_i^{(m)}).
\end{equation}

The final fused spot-level representation is a weighted sum over experts:
\begin{equation}
z_i = \sum_{m} s_i^{(m)} \cdot h_i^{(m)}.
\end{equation}

In rare cases where all \(\beta_i^{(m)} < \gamma\), the fused representation defaults to the expert output of the modality with the highest original confidence:
\begin{equation}
z_i = h_i^{(m^*)}, \quad m^* = \arg\max_{m} \beta_i^{(m)}.
\end{equation}

Finally, we collect the fused representations \( z_i \) of all spots into a set, denoted as
\begin{equation}
Z = \{ z_1, z_2, \ldots, z_N \},
\end{equation}
which forms a unified representation under the spatial structure \( S \), corresponding to the output of Equation~\eqref{eq:Z}.

This unified representation \( Z \) comprehensively captures the integrated features of multimodal spatial omics data, facilitating subsequent downstream analysis tasks.

To reconstruct modality-specific features from the fused embeddings, we employ a graph-based decoder that utilizes the spatial adjacency structure \( \hat{A} \). Specifically, for each modality \( m \in \{R, A, I\} \), we define the reconstructed features as:
\begin{equation}
\hat{F}^{(m)} = \sigma\left( \hat{A} \cdot \mathcal{F}^{(m)}(Z) \right),
\label{eq:gcn-decoder}
\end{equation} 

This formulation ensures that reconstruction leverages both the fused representation and the spatial context encoded in \( \hat{A} \).

The reconstruction loss for each modality is computed as:
\begin{equation}
\mathcal{L}_{\mathrm{rec}}^{(m)} = \frac{1}{N} \sum_{i=1}^{N} \left\| f_i^{(m)} - \hat{f}_i^{(m)} \right\|_2^2,
\end{equation}

where \( \hat{f}_i^{(m)} \) is the reconstructed feature for spot \( i \) in modality \( m \), obtained from \( \hat{F}^{(m)} \).

The overall training objective is formulated as:
\begin{equation}
\mathcal{L}_{\mathrm{total}} = \sum_{m \in \{R, A, I\}} \mathcal{L}_{\mathrm{rec}}^{(m)} + \lambda \sum_{m_i \ne m_j} \mathcal{L}_{\mathrm{contrast}}^{m_i, m_j},
\end{equation}

where \( \lambda \) is a hyperparameter balancing the contribution of the contrastive alignment loss. In our experiments, we set \( \lambda = 2 \).

\begin{algorithm}[H]
\caption{GROVER}
\label{alg:grover}
\textbf{Input:} Multi-modal features $\{F^{(m)}\}_{m=1}^M$, spatial coordinates $S = \{(x_i, y_i)\}_{i=1}^N$ \\
\textbf{Parameters:} Epochs $E$, temperature $\tau$, contrastive loss weight $\lambda$, confidence threshold $\gamma$
\begin{algorithmic}[1]
\STATE Construct graphs: $\mathcal{G}_S$ from $S$ and $\mathcal{G}_F^{(m)}$ from $F^{(m)}$ via KNN
\FOR{$e = 1$ to $E$}
    \FOR{each modality $m$}
        \STATE Extract $e_i^S$ and $e_i^{F^{(m)}}$ via KAN-GCN (Eq.~8)
        \STATE Fuse to obtain $\tilde{e}_i^{(m)}$ via attention (Eq.~11)
    \ENDFOR
    \STATE Contrastive loss: $\mathcal{L}_{\text{contrast}}$ across $\{\tilde{e}_i^{(m)}\}$ (Eq.~16)
    \STATE MoE routing to get $Z$ (Eq.~22, 24)
    \STATE Reconstruction loss: $\mathcal{L}_{\text{recon}}$ from $Z$ (Eq.~26)
    \STATE Update model via total loss $\mathcal{L}_{\text{total}}$ (Eq.~27)
\ENDFOR
\end{algorithmic}
\textbf{Output:} Unified comprehensive embeddings $Z$
\end{algorithm}
\vspace{-1em}
\noindent
The resulting embeddings $Z$ serve as comprehensive representations for downstream analysis tasks, such as clustering and spatial domain detection.
\vspace{-0.3em}

\section{Experiments}
\subsection{Exprimental Setups}

\begin{table*}[t]
\centering
\small
\setlength\tabcolsep{5pt}
\renewcommand{\arraystretch}{0.9}
\label{tab:multi-dataset-results}
\begin{tabular}{lccccccccc}
\toprule
      Method & ARI (\%↑) & NMI (\%↑) & FMI (\%↑) & SC (\%↑) & AMI (\%↑) & Jaccard (\%↑) & CHI (↑) & Purity (\%↑) & DBI (\%↓) \\
      \midrule
      \multicolumn{10}{c}{Human Tonsil dataset} \\
      \midrule
      GROVER   & \textbf{45.2±7.8} & \textbf{54.3±9.9} & \textbf{54.1±6.8} & \textbf{31.6±3.9} & \textbf{54.2±10.1} & \textbf{37.3±6.6} & \textbf{2494.4±285.5} & \textbf{69.4±5.4} & \textbf{139.8±10.5} \\
      MISO     & 41.3±6.7 & 51.2±4.6 & \underline{52.5±4.3} & 7.0±1.6 & 51.2±4.6 & \underline{35.4±3.8} & 244.4±14.6 & 64.2±5.5 & 203.4±14.8 \\
      Spatialglue & \underline{43.3±6.7} & \underline{53.9±8.9} & 52.4±6.1 & \underline{23.8±3.2} & \underline{53.9±8.9} & 35.3±5.6 & \underline{1063.6±123.6} & \underline{68.7±5.0} & 159.6±7.0 \\
      COSMOS   & 19.8±6.7 & 27.9±6.0 & 32.3±6.6 & 20.0±0.7 & 27.6±5.6 & 19.3±4.9 & 937.4±99.6 & 49.9±9.0 & \underline{157.8±4.2} \\
      \midrule
      \multicolumn{10}{c}{Human Breast Cancer} \\
      \midrule
      GROVER   & \textbf{44.1±10.7} & \underline{52.4±8.7} & \textbf{53.9±8.6} & \textbf{36.3±7.7} & \underline{52.3±8.6} & \textbf{37.3±8.1} & \textbf{2436.3±385.1} & \underline{64.8±9.9} & \textbf{139.6±13.8} \\
      MISO     & 37.5±3.0 & 47.9±2.0 & 49.8±3.0 & 11.0±0.6 & 47.7±2.0 & 32.7±2.7 & 289.4±20.8 & 56.7±3.6 & 211.5±10.7 \\
      Spatialglue & \underline{43.0±6.9} & \textbf{53.0±5.1} & \underline{52.1±6.1} & 20.2±0.8 & \textbf{53.5±4.8} & \underline{35.2±6.0} & 1175.0±135.1 & \textbf{67.2±5.0} & 172.2±3.3 \\
      COSMOS   & 25.6±2.2 & 36.5±3.5 & 37.0±1.8 & \underline{24.8±0.8} & 36.3±3.5 & 22.7±1.6 & \underline{1226.4±106.1} & 54.5±2.9 & \underline{143.4±2.6} \\
      \midrule
      \multicolumn{10}{c}{Human Glioblastoma} \\
      \midrule
      GROVER   & \underline{40.8±6.6} & \textbf{53.9±4.1} & \underline{51.6±4.6} & 22.6±1.1 & \textbf{53.8±3.8} & \underline{34.1±4.8} & \underline{1412.9±110.7} & \textbf{71.9±3.1} & \underline{157.0±3.8} \\
      MISO     & \textbf{43.5±6.9} & 49.2±2.2 & \textbf{55.5±7.0} & 9.6±2.9 & 49.0±2.2 & \textbf{38.4±7.2} & 421.2±47.1 & 65.3±7.5 & 235.8±10.3 \\
      Spatialglue & 40.1±7.6 & \underline{53.8±7.3} & 50.9±5.5 & \underline{23.4±0.5} & \underline{53.8±7.3} & 33.4±5.4 & \textbf{1430.3±132.7} & \underline{72.3±3.6} & 157.2±3.9 \\
      COSMOS   & 32.0±6.9 & 48.6±4.3 & 44.2±4.5 & \textbf{25.8±2.4} & 48.4±4.2 & 28.0±4.1 & 1325.2±92.3 & 67.8±3.9 & \textbf{137.4±9.0} \\
      \midrule
      \multicolumn{10}{c}{Human Tonsil with Add-on Antibodies} \\
      \midrule
      GROVER   & \textbf{46.5±5.6} & \textbf{59.0±4.8} & \underline{55.3±6.0} & \textbf{38.2±1.2} & \textbf{58.8±4.7} & \underline{38.0±5.7} & \textbf{3979.0±185.4} & \textbf{70.5±6.1} & \textbf{105.8±2.9} \\
      MISO     & 44.6±11.9 & 56.1±7.6 & \textbf{55.9±10.4} & 8.3±0.5 & 55.9±7.6 & \textbf{38.9±10.2} & 356.7±32.5 & 65.5±11.0 & 217.2±15.2 \\
      Spatialglue & \underline{45.3±7.3} & \underline{58.1±5.7} & 54.1±7.3 & \underline{21.4±1.1} & \underline{58.0±5.8} & 36.9±6.6 & \underline{1331.3±133.9} & \textbf{70.5±5.9} & \underline{160.6±2.9} \\
      COSMOS   & 24.6±4.3 & 35.1±1.0 & 36.4±5.7 & 18.4±2.5 & 35.0±0.9 & 22.1±4.2 & 1194.0±139.2 & 51.5±6.8 & 168.8±5.7 \\
      \bottomrule
    \end{tabular}%
\caption{
Performance comparison of GROVER and baselines on four spatial multi-omics datasets using nine clustering metrics. Bold indicates the best result, and underline denotes the second-best. All metrics are the higher the better, except for DBI, where a lower value indicates better performance.
}
\label{tab:multi-dataset-results}
\end{table*}

\begin{table*}[t]
\centering
\small
\setlength\tabcolsep{4pt}
\renewcommand{\arraystretch}{0.9}
\label{tab:ablation}
\begin{tabular}{lccccccccc}
\toprule
Method & ARI (\%↑) & NMI (\%↑) & FMI (\%↑) & SC (\%↑) & CHI (↑) & Purity (\%↑) & AMI (\%↑) & Jaccard (\%↑) & DBI (\%↓) \\
\midrule
GROVER           
& \textbf{46.5±5.6} 
& \textbf{59.0±4.8} 
& \textbf{55.3±6.0} 
& 38.2±1.2
& 3979.0±185.4 
& \textbf{70.5±6.1} 
& \textbf{58.8±4.7} 
& \textbf{38.0±5.7} 
& 105.8±2.9 \\
w/o MoE          
& 42.5±4.3          
& 56.8±3.0          
& 51.6±4.8          
& 21.8±1.2          
& 1081.5±94.8           
& 69.9±4.3          
& 56.7±3.1          
& 34.6±4.5          
& 158.6±5.9 \\
w/o $\mathcal{L}_{\mathrm{contrast}}$      
& 45.5±7.2          
& 57.8±4.3          
& 54.1±7.4          
& 21.6±2.6          
& 2201.3±306.1          
& 68.4±6.1          
& 57.7±4.3          
& 37.3±7.2          
& 168.8±10.5 \\
w/o KAN-GCN      
& 42.7±6.7          
& 55.9±5.2          
& 52.5±6.7          
& \textbf{52.6±1.1}          
& \textbf{6708.4±247.3}          
& 68.7±6.7          
& 55.9±5.2          
& 35.6±6.3          
& \textbf{89.6±4.7} \\
\bottomrule
\end{tabular}
\caption{Ablation study results on the Human Tonsil with add-on antibodies (results on other datasets are provided in the Appendix). GROVER achieves strong performance on most clustering metrics. Removing the self-adaptive MoE module, contrastive loss, or the KAN-GCN leads to noticeable performance degradation, confirming the effectiveness of each component.}
\label{tab:ablation}
\vspace{-5mm}
\end{table*}

\subsubsection{Dataset.}We conduct quantitative and qualitative experiments on four public datasets to verify the effectiveness of the proposed method: 1) 10x Visium human breast Cancer gene and protein expression dataset \cite{10x_breast_2023}; 2) 10x Visium human glioblastoma gene and protein expression dataset \cite{10x_gbm_2023}; 3) 10x Visium human tonsil gene and protein expression dataset \cite{10x_tonsil_2_2023}; 4) 10x Visium human tonsil with add-on antibodies gene and protein expression dataset \cite{10x_tonsil_2023}; These datasetsare detailed in the Appendix\textsuperscript{1}
.

\subsubsection{Baselines.}We compare our approach with three recent state-of-the-art multimodal omics methods: MISO, SpatialGlue, and COSMOS. Notably, MISO supports image modality integration.

\subsubsection{Metrics.} We selected nine diverse metrics to comprehensively evaluate the model performance, including Adjusted Rand Index (ARI) \cite{steinley2004properties}, Normalized Mutual Information (NMI) \cite{vinh2009information}, Fowlkes–Mallows Index (FMI) \cite{fowlkes1983method}, Silhouette Coefficient (SC) \cite{rousseeuw1987silhouettes}, Adjusted Mutual Information (AMI) \cite{vinh2009information}, Jaccard Index \cite{niwattanakul2013using}, Davies–Bouldin Index (DBI) \cite{davies1979cluster}, Calinski–Harabasz Index (CHI) \cite{calinski1974dendrite}, and Purity. Detailed experimental settings are provided in the Appendix.

\subsubsection{Resources.}
All experiments were run on a workstation with dual NVIDIA RTX A5000 GPUs (24 GB) and dual Intel Xeon Silver 4210R CPUs (2.40 GHz, 20 cores × 2). GROVER converged within 300 epochs under this setup.

\vspace{-0.3em}
\subsection{Quantitative Experimental Results}
We conducted quantitative experiments on four real-world spatial multi-omics datasets. Due to the absence of ground-truth annotations spanning all modalities, we evaluated performance using cell-type clustering labels derived separately from RNA and ADT, and reported the mean and standard deviation across five clustering settings (10 to 6). Since RNA and ADT labels reflect related but non-identical biological groupings, certain methods show slightly higher variance on specific datasets. As shown in Table~\ref{tab:multi-dataset-results}, GROVER consistently ranks among the top-performing methods across nine clustering metrics and all four datasets. For example, it improves ARI by 4.4\% and SC by 32.8\% on the Human Tonsil dataset, and boosts ARI and FMI by 2.6 and 3.3 percentage points on Human Breast Cancer. These results demonstrate GROVER’s strong ability to integrate heterogeneous modalities while preserving spatial and structural coherence. Notably, the bimodal method SpatialGlue often outperforms the trimodal method MISO, suggesting that uniform fusion of all modalities may not be ideal. In contrast, GROVER's adaptive fusion strategy dynamically adjusts modality weights per spot, leading to more effective integration tailored to local data characteristics.

\vspace{-0.3em}
\subsection{Qualitative Experimental Results}
We performed qualitative clustering on four spatial multi-omics datasets with histological images. Taking the Human Tonsil dataset as an example (Figure 2), GROVER’s fusion better reconstructs biological structures and clearly outlines germinal center (GC) regions and boundaries. Unlike MISO, which equally fuses all modalities, GROVER selectively emphasizes informative ones to identify regions requiring multi-modal synergy. While SpatialGlue also detects GC regions, it often splits the same type into subgroups, revealing the limitations of bi-modal integration.

\begin{figure*}[t]
  \centering
  \includegraphics[width=0.9\textwidth]{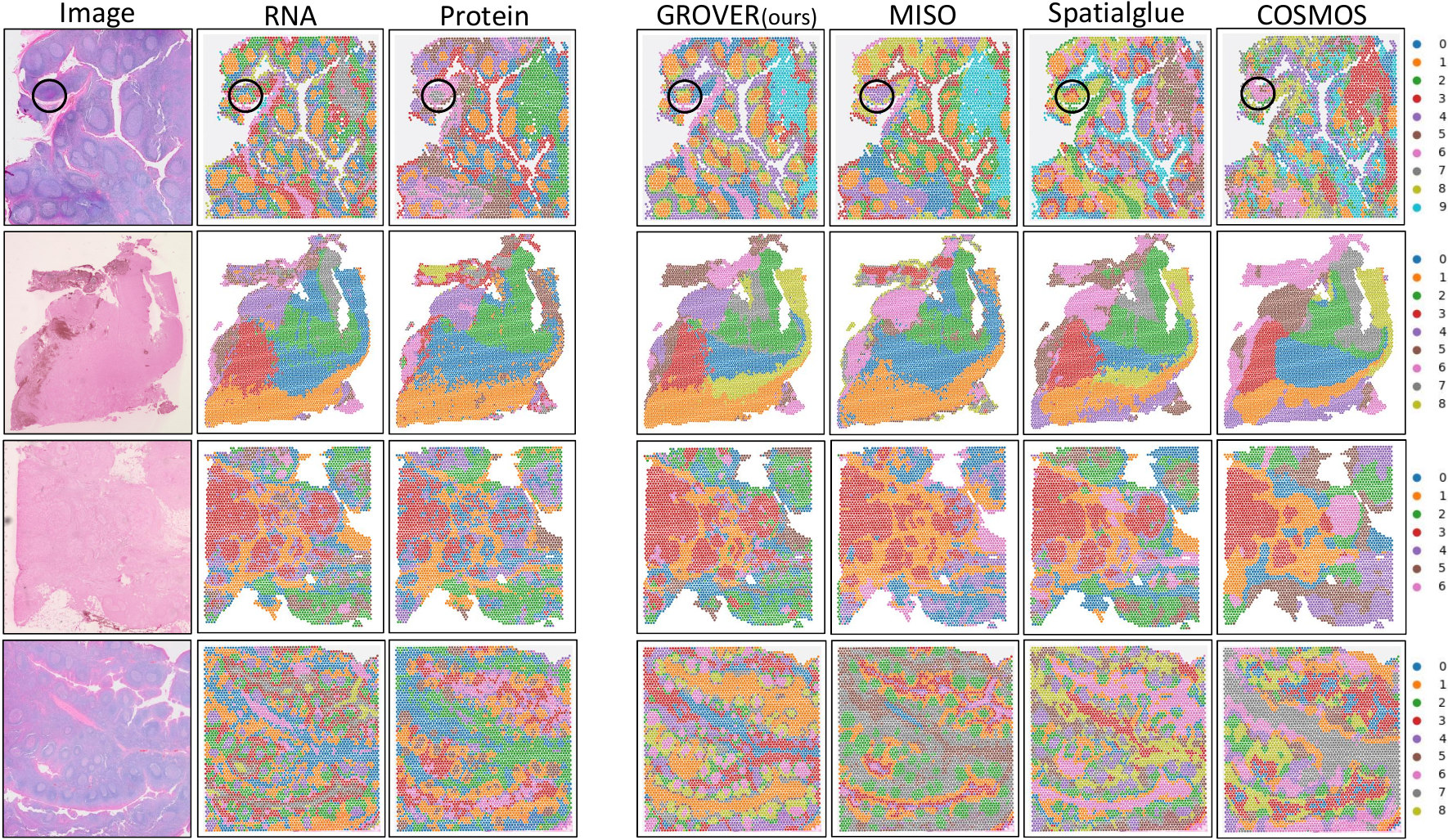}
  \caption{Visualization of clustering results by GROVER and baseline methods on four spatial multi-omics datasets. From top to bottom: (1) Human Tonsil, (2) Human Glioblastoma, (3) Human Breast Cancer, and (4) Human Tonsil with Add-on Antibodie.}
  \label{fig:grover}
  \vspace{-0.8em} 
\end{figure*}

\subsection{Ablation Studies}
To evaluate the effectiveness of each core component in the GROVER framework, we performed ablation studies on the Human Tonsil dataset with add-on antibodies. As shown in Table~\ref{tab:ablation}, removing the expert routing in the MoE module and replacing it with a simple summation of modality-specific features led to notable performance drops, with ARI, NMI, and FMI decreasing by 4.0\%, 2.2\%, and 3.7\%, respectively. This highlights the importance of dynamically assigning experts to handle heterogeneous signal quality. Removing the spot-feature-pair contrastive loss reduced SC by 16.6\%, showing its role in preserving spatial coherence. Replacing the KAN-GCN encoder with a standard GCN slightly improved unsupervised metrics SC and DBI, but clearly decreased supervised metrics, demonstrating the advantage of KAN's nonlinear modeling in capturing complex spatial-feature interactions. Together, these results validate the necessity and effectiveness of the expert routing strategy, contrastive learning design, and KAN-GCN based encoder within the overall GROVER architecture.

\subsection{Parameter Sensitivity Experiments}
\begin{figure}[t]
  \centering
    \includegraphics[width=1\linewidth]{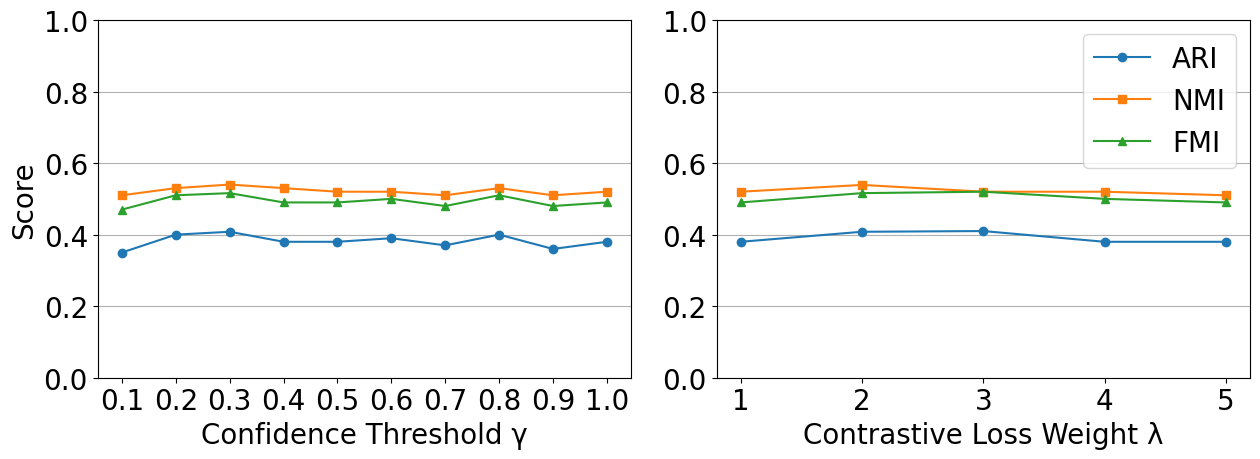}
    \caption{Parameter sensitivity analysis of GROVER on the Human Glioblastoma dataset.}
    \label{fig:sensitivity}
\vspace{-1.5em}
\end{figure}

We conducted parameter sensitivity experiments to evaluate GROVER’s robustness to key hyperparameters, including the confidence threshold $\gamma$ and contrastive loss weight $\lambda$. As shown in Figure~\ref{fig:sensitivity}, GROVER exhibits stable performance across a range of $\gamma$ values in terms of ARI, NMI, and FMI, indicating that the model is not particularly sensitive to this parameter.

Specifically, extremely low values such as $\gamma = 0.1$ lead to suboptimal results, as the model fails to filter out unreliable modalities. Performance peaks around $\gamma = 0.2$ and $0.3$, where most spots with normal-quality signals are able to incorporate all modalities and benefit from richer information. Although performance remains relatively stable as $\gamma$ increases, it slightly declines beyond this range due to over-reliance on a single modality, causing loss of complementary features. Based on these observations, we set $\gamma = 0.3$ in our experiments, as it lies near the reciprocal of the number of experts and strikes a good balance between modality filtering and integration.In contrast, the model demonstrates even greater robustness to the contrastive loss weight $\lambda$, with minimal variation observed across metrics, confirming the stability of the contrastive learning component.

\section{Conclusion}
In this paper, we propose GROVER, an adaptive framework that integrates spatial multi-omics data at single-spot resolution using spatial and modality-specific graphs. With a multi-expert fusion mechanism and spot-feature-pair contrastive learning, GROVER dynamically routes information based on data quality, enabling robust integration under high heterogeneity or noise. It achieves superior performance across multiple metrics on four benchmark datasets.

\section{Acknowledgements}
This work was supported by National Natural Science Foundation of China (32300554). The computational resources are supported by SongShan Lake HPC Center (SSL-HPC) in Great Bay University.

\bibliography{aaai2026}

\end{document}